\documentclass[conference,10pt,letterpaper]{IEEEtran}% 
\usepackage{amsmath}% for double integral symbol in this template
\usepackage{graphicx}% for figures
\usepackage{multirow}% to allow multiple-row elements in tabular environment
\usepackage[none]{hyphenat}% turn off hyphenation to make text extraction and indexing easier
\usepackage{float}% better control of floating figures and tables
\usepackage{subfig}% for subfigures within figures
\usepackage{dblfloatfix}% fix to allow page-wide floats at bottom of page
\usepackage{tikz}
\usepackage{t1enc}% allows access to various special characters
\usepackage{times}% change font to Nimbus Roman (based on Times Roman)
\usepackage{siunitx}
\usepackage{cite}
\usepackage{balance} 
\usepackage{hyperref} 

\makeatletter

\def\@IMSauthorblockNAMEstyle{\normalfont\IMSauthorsize}
\def\@IMSauthorblockAFFILstyle{\normalfont\IMSaffilsize}
\def\@IMSauthorblockEMAILstyle{\normalfont\IMSaffilsize}
\def\IMSauthorblockNAME#1{%
\relax\@IMSauthorblockNAMEstyle%
#1%
}%
\def\IMSauthorblockAFFIL#1{%
\relax\@IMSauthorblockAFFILstyle%
\vskip\@IEEEauthorblockAtopspace
#1%
}%
\def\IMSauthorblockEMAIL#1{%
\relax\@IMSauthorblockEMAILstyle%
\vskip\@IEEEauthorblockAtopspace
#1%
}%
\newcommand{\IMSauthor}[1]{%
\ifIsBlindReviewVersion%
\author{\phantom{\parbox{\textwidth}{\center\relax#1}}}%
\else%
\author{\parbox{\textwidth}{\center\relax#1}}%
\fi%
}%
\newif\ifIsBlindReviewVersion
\def\IMSthispaperforblindreview{\IsBlindReviewVersiontrue}
\def\IMSthispaperforfinalpublication{\IsBlindReviewVersionfalse}
\def\@maketitle{\newpage
\bgroup\par\addvspace{0.5\baselineskip}\centering%
\ifCLASSOPTIONtechnote% technotes
   {\bfseries\large\@IEEEcompsoconly{\sffamily}\@title\par}\vskip 1.3em{\lineskip .5em\@IEEEcompsoconly{\sffamily}\@author
   \@IEEEspecialpapernotice\par{\@IEEEcompsoconly{\vskip 1.5em\relax
   \@IEEEtitleabstractindextextbox{\@IEEEtitleabstractindextext}\par
   \hfill\@IEEEcompsocdiamondline\hfill\hbox{}\par}}}\relax
\else% not a technote
   \vskip0.2em{\IMStitlesize\ifCLASSOPTIONtransmag\bfseries\LARGE\fi\@IEEEcompsoconly{\sffamily}\@IEEEcompsocconfonly{\normalfont\normalsize\vskip 2\@IEEEnormalsizeunitybaselineskip
   \bfseries\Large}\@title\par}\vskip1.0em\par% CAUSAL PRODUCTIONS change on this line
   \ifCLASSOPTIONconference%
      {\@IEEEspecialpapernotice\mbox{}\vskip\@IEEEauthorblockconfadjspace%
       \mbox{}\hfill\begin{@IEEEauthorhalign}\@author\end{@IEEEauthorhalign}\hfill\mbox{}\par}\relax
   \else% peerreviewca, peerreview or journal
      \ifCLASSOPTIONpeerreviewca
         {\@IEEEcompsoconly{\sffamily}\@IEEEspecialpapernotice\mbox{}\vskip\@IEEEauthorblockconfadjspace%
          \mbox{}\hfill\begin{@IEEEauthorhalign}\@author\end{@IEEEauthorhalign}\hfill\mbox{}\par
          {\@IEEEcompsoconly{\vskip 1.5em\relax
           \@IEEEtitleabstractindextextbox{\@IEEEtitleabstractindextext}\par\hfill
           \@IEEEcompsocdiamondline\hfill\hbox{}\par}}}\relax
      \else% journal, peerreview or transmag
         \ifCLASSOPTIONtransmag
           {\@IEEEspecialpapernotice\mbox{}\vskip\@IEEEauthorblockconfadjspace%
            \mbox{}\hfill\begin{@IEEEauthorhalign}\@author\end{@IEEEauthorhalign}\hfill\mbox{}\par
           {\vspace{0.5\baselineskip}\relax\@IEEEtitleabstractindextextbox{\@IEEEtitleabstractindextext}\vspace{-1\baselineskip}\par}}\relax
         \else% journal or peerreview
           {\lineskip.5em\@IEEEcompsoconly{\sffamily}\sublargesize\@author\@IEEEspecialpapernotice\par
           {\@IEEEcompsoconly{\vskip 1.5em\relax
            \@IEEEtitleabstractindextextbox{\@IEEEtitleabstractindextext}\par\hfill
            \@IEEEcompsocdiamondline\hfill\hbox{}\par}}}\relax
         \fi
      \fi
   \fi
\fi\par\addvspace{0.0\baselineskip}\egroup}% CAUSAL PRODUCTIONS change on this line, reduce the vspace from 0.5\baselineskip to 0.0

\def\IMStitlesize{\@setfontsize{\IMStitlesize}{18}{21pt}}% CAUSAL PRODUCTIONS change on this line
\def\IMSauthorsize{\@setfontsize{\IMSauthorsize}{12}{13pt}}% CAUSAL PRODUCTIONS change on this line
\def\IMSaffilsize{\@setfontsize{\IMSaffilsize}{12}{13pt}}% CAUSAL PRODUCTIONS change on this line
\def\IMScaptionsize{\@setfontsize{\IMScaptionsize}{8}{9pt}}% CAUSAL PRODUCTIONS change on this line
\def\IMSbibsize{\@setfontsize{\IMSbibsize}{8}{9pt}}% CAUSAL PRODUCTIONS change on this line

\def\@IEEEauthorblockNstyle{\IMSauthorsize\@IEEEcompsocnotconfonly{\sffamily}\@IEEEcompsocconfonly{\large}}%CAUSAL PRODUCTIONS removed sublargesize to get correct IMSauthorsize
\def\@IEEEauthorblockAstyle{\IMSaffilsize\@IEEEcompsocnotconfonly{\sffamily}\@IEEEcompsocconfonly{\itshape}\@IEEEcompsocconfonly{\large}}%CAUSAL PRODUCTIONS removed normalsize to get correct IMSaffilsize
\def\@IEEEauthordefaulttextstyle{\IMSauthorsize\@IEEEcompsocnotconfonly{\sffamily}\sublargesize}%CAUSAL PRODUCTIONS

\def\thebibliography#1{\section*{\refname}%
    \addcontentsline{toc}{section}{\refname}%
    \IMSbibsize\@IEEEcompsocconfonly{\small}\vskip 0.3\baselineskip plus 0.1\baselineskip minus 0.1\baselineskip% CAUSAL PRODUCTIONS change on this line
    \list{\@biblabel{\@arabic\c@enumiv}}%
    {\settowidth\labelwidth{\@biblabel{#1}}%
    \leftmargin\labelwidth
    \advance\leftmargin\labelsep\relax
    \itemsep \IEEEbibitemsep\relax
    \usecounter{enumiv}%
    \let\p@enumiv\@empty
    \renewcommand\theenumiv{\@arabic\c@enumiv}}%
    \let\@IEEElatexbibitem\bibitem%
    \def\bibitem{\@IEEEbibitemprefix\@IEEElatexbibitem}%
\def\newblock{\hskip .11em plus .33em minus .07em}%
\ifCLASSOPTIONtechnote\sloppy\clubpenalty4000\widowpenalty4000\interlinepenalty100%
\else\sloppy\clubpenalty4000\widowpenalty4000\interlinepenalty500\fi%
    \sfcode`\.=1000\relax}

\long\def\@makecaption#1#2{%
\ifx\@captype\@IEEEtablestring%
\par\@IEEEtabletopskipstrut% strut used to align table caption with facing column
\else
\@IEEEfigurecaptionsepspace
\fi
\setbox\@tempboxa\hbox{\normalfont\IMScaptionsize {#1.}\nobreakspace\nobreakspace #2}%
\ifdim \wd\@tempboxa >\hsize%
\setbox\@tempboxa\hbox{\normalfont\IMScaptionsize {#1.}\nobreakspace\nobreakspace}%
\parbox[t]{\hsize}{\normalfont\IMScaptionsize\noindent\unhbox\@tempboxa#2}%
\else
\ifCLASSOPTIONconference \hbox to\hsize{\normalfont\IMScaptionsize\hfil\box\@tempboxa\hfil}%
\else \hbox to\hsize{\normalfont\IMScaptionsize\box\@tempboxa\hfil}%
\fi\fi
\ifx\@captype\@IEEEtablestring%
\@IEEEtablecaptionsepspace
\else
\fi}

\newlength\tablecaptiontotableskip
\newlength\figuretocaptionskip
\def\@IEEEfigurecaptionsepspace{\vskip\figuretocaptionskip\relax}%
\def\@IEEEtablecaptionsepspace{\vskip\tablecaptiontotableskip\relax}%

\def\abstract{\normalfont%
\@IEEEabskeysecsize\bfseries\textit{\abstractname}\,\bfseries\textit{---}\,%
\@IEEEgobbleleadPARNLSP}%

\def\IEEEkeywords{\normalfont%
\@IEEEabskeysecsize\bfseries\textit{\IEEEkeywordsname}\,\bfseries\textit{---}\,%
\@IEEEgobbleleadPARNLSP}%
\def\endIEEEkeywords{\relax\vspace{0.67ex}%
\par\if@twocolumn\else\endquotation\fi%
\normalsize\normalfont}%

\DeclareRobustCommand*{\IMSauthorrefmark}[1]{\raisebox{0pt}[0pt][0pt]{\textsuperscript{\footnotesize{#1}}}}%
\def\@IEEEauthorblockNtopspace{0ex}
\def\@IEEEauthorblockAtopspace{1mm}
\def\IEEEkeywordsname{Keywords}% use Keywords instead of Index Terms
\def\subsubsection{\@startsection{subsubsection}{3}{\z@}{1.5ex plus 1.5ex minus 0.5ex}%
{0.7ex plus .5ex minus 0ex}{\normalfont\normalsize\itshape}}%
\def\@seccntformat#1{\csname the#1dis\endcsname\relax}% moved the spacer \hskip 0.5em to individual handlers below
\def\thesubsectiondis{{\hbox to\parindent{\Alph{subsection}.}}}%		B.	% CAUSAL PRODUCTIONS: indent the subsection name to match paragraph indent
\def\thesubsubsectiondis{{\hbox to \parindent{\arabic{subsubsection})}}}%	3)	% CAUSAL PRODUCTIONS: indent the subsubsection name to match paragraph indent
\def\theparagraphdis{{\hbox to \parindent{\alph{paragraph})}}}%			d)	% CAUSAL PRODUCTIONS: indent the subsubsubsection name to match paragraph indent
\IEEEilabelindentA \parindent
\IEEEilabelindent \IEEEilabelindentA
\IEEEelabelindent \parindent
\IEEEdlabelindent \parindent
\IEEElabelindent \parindent
\newlength\@IMSparindent
\newcommand\IMSdisplayacksection[1]{%
\ifIsBlindReviewVersion%
\noindent\phantom{\parbox[t]{\columnwidth}{\normalbaselines\setlength{\parindent}{\@IMSparindent}{#1}\strut}}%\IMSacktext
\else%
\noindent\parbox[t]{\columnwidth}{\normalbaselines\setlength{\parindent}{\@IMSparindent}{#1}\strut}%
\fi%
}%

\makeatother

\newcommand\copyrighttext{%
  \footnotesize \textcopyright 2024 IEEE. Personal use of this material is permitted.
  Permission from IEEE must be obtained for all other uses, in any current or future
  media, including reprinting/republishing this material for advertising or promotional
  purposes, creating new collective works, for resale or redistribution to servers or
  lists, or reuse of any copyrighted component of this work in other works.}
\newcommand\copyrightnotice{%
\begin{tikzpicture}[remember picture,overlay]
\node[anchor=south,yshift=10pt] at (current page.south) {\fbox{\parbox{\dimexpr\textwidth-\fboxsep-\fboxrule\relax}{\copyrighttext}}};
\end{tikzpicture}%
}

\begin{document}
%%%%%%%%%%%%%%%%%%%%%%%%%%%%%%%%%%%%%%%%%%%%%%%%%%%%%%%%%%%%%%%%%%%%%%%%%%%%%
% We use \raggedbottom to avoid latex adding vertical space around headings.
% This gives a better idea to the author about how much white space remains
% as the page limit is approached.
\raggedbottom
%
%%%%%%%%%%%%%%%%%%%%%%%%%%%%%%%%%%%%%%%%%%%%%%%%%%%%%%%%%%%%%%%%%%%%%%%%%%%%%
% PAPER TITLE AND AUTHOR BLOCK
%
% The paper title can use linebreaks \\ within to get better formatting if desired.
%
\title{Radar-Based Recognition of Static Hand Gestures\\in American Sign Language}
%
% Next we define the author names and affiliations.
% Author names are listed using \IMSauthorblockNAME{} with comma separators between names.
% Affiliations are listed using \IMSauthorblockAFFIL{} with \\ separators between affiliations.
% Email addresses are listed using \IMSauthorblockEMAIL{} with comma separators between emails.
% See below for examples of each of these.
%
% Symbols marking author-affiliation relations are output using \IMSauthorrefmark{}.
%
% Next we typeset the authorblock either as visible text, or as an empty
% box of the same size, based on the value of the Blind Review Flag.
% Note that the Blind Review Flag also determines whether the Acknowledgements
% section is visible or invisible.
% To set the flag to Blind Review mode, simply uncomment the next line
\IMSthispaperforblindreview
% or to set the flag to Final Paper mode (with author block visible) then
% simply uncomment the next line:
\IMSthispaperforfinalpublication
\IMSauthor{%
\IMSauthorblockNAME{% Author Names
Christian Schuessler\IMSauthorrefmark{\#1*},
Wenxuan Zhang\IMSauthorrefmark{\#*},
Johanna Bräunig\IMSauthorrefmark{\#*},
Marcel Hoffmann\IMSauthorrefmark{\#*},
Michael Stelzig\IMSauthorrefmark{\#*},
Martin Vossiek\IMSauthorrefmark{\#*}
}% end of \IMSauthorblockNAME
\\%
\IMSauthorblockAFFIL{% Author Affiliations
\IMSauthorrefmark{\#}Institute of Microwaves and Photonics\\
\IMSauthorrefmark{*}Friedrich-Alexander-Universität Erlangen-Nürnberg, Erlangen, Germany
}% end of \IMSauthorblockAFFIL
\\%
\IMSauthorblockEMAIL{% Author Emails
\IMSauthorrefmark{1}christian.schuessler@fau.de
}% end of \IMSauthorblockEMAIL
}% end of \IMSauthor
%
% Next we make the title/author block using the information defined above.
\maketitle
\copyrightnotice
%
%%%%%%%%%%%%%%%%%%%%%%%%%%%%%%%%%%%%%%%%%%%%%%%%%%%%%%%%%%%%%%%%%%%%%%%%%%%%%
% ABSTRACT paragraph.
% in the abstract paragraph.
%
\begin{abstract}

% my rought version
%With the rise of virtual reality (VR), automatic gesture recognition becomes increasingly important. Especially, 
%the automatic recognition of hand signs is of great importance, 
%since it offers an easy way to control and navigate through VR applications. 
%Compared to cameras, radar sensors offer more privacy, since they do not measure human conceivable details. 
%This is due to their lower resolution and different wavelength compared to visual light.
%Compared to most works, which employ radar sensors for dynamic hand gesture recognition by exploiting Doppler information,
%this approach focuses on the classification with an imaging radar working on spatial information.
%Since the generation of sufficient training data is time consuming and may not include all edge cases, this work investigates
%the usage of synthetic data generated by a radar ray tracing simulator, that already showed profiency in simulating 
%microwave images of hand poses.
%Even the trained neural network (NN) was solely trained on syntehtic data it still shows promising evaluated on real measurement data. 

% improved by chat gpt and with some extension from me
In the fast-paced field of human-computer interaction (HCI) and virtual reality (VR), automatic gesture recognition has become increasingly essential. This is particularly true for the recognition of hand signs, providing an intuitive way to effortlessly navigate and control VR and HCI applications. Considering increased privacy requirements, radar sensors emerge as a compelling alternative to cameras. They operate effectively in low-light conditions without capturing identifiable human details, thanks to their lower resolution and distinct wavelength compared to visible light.

While previous works predominantly deploy radar sensors for dynamic hand gesture recognition based on Doppler information, our approach prioritizes classification using an imaging radar that operates on spatial information, e.g. image-like data.
However, generating large training datasets required for neural networks (NN) is a time-consuming and challenging process, often falling short of covering all potential scenarios. Acknowledging these challenges, this study explores the efficacy of synthetic data generated by an advanced radar ray-tracing simulator. This simulator employs an intuitive material model that can be adjusted to introduce data diversity.

Despite exclusively training the NN on synthetic data, it demonstrates promising performance when put to the test with real measurement data. This emphasizes the practicality of our methodology in overcoming data scarcity challenges and advancing the field of automatic gesture recognition in VR and HCI applications.
\end{abstract}
%
%%%%%%%%%%%%%%%%%%%%%%%%%%%%%%%%%%%%%%%%%%%%%%%%%%%%%%%%%%%%%%%%%%%%%%%%%%%%%
% THE REST OF THE PAPER follows.
%

\section{Introduction}
\begin{figure}[t]
    \centering
    \includegraphics[width=\columnwidth]{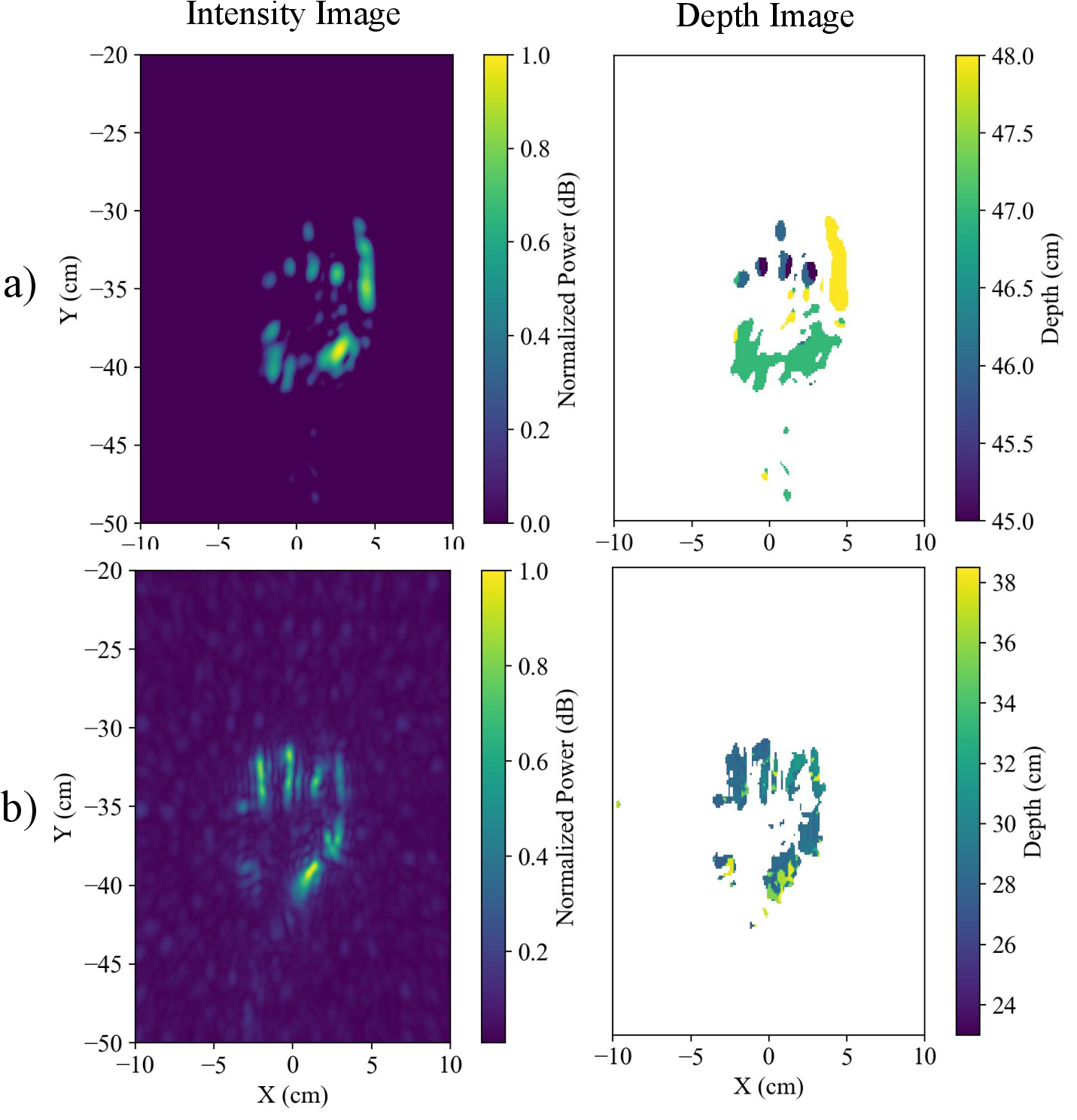}
    \caption{Example showing the hand sign 'A' from ASL. In the first row (a), the intensity and depth image of the simulation are shown. In the second row (b), the same images are shown for a real measurement.}
    \label{fig:firstImage}
\end{figure}

In the emerging field of virtual reality (VR) or human computer interaction (HCI), in general, automatic hand gesture recognition allows for an easy and intuitive control.
Nowadays, most hand gesture recognition systems are built upon RGB cameras~\cite{lit:FangRealTimeHand, lit:ReviewHandGestureRecognition}, or RGB-D cameras~\cite{lit:ReviewHandGestureDepth}.
The advantage of these systems is that they can be built upon relatively inexpensive sensors, and the field of computer vision is extremely well studied.
However, especially for privacy-sensitive environments, harsh conditions, and low-light conditions, radar sensors are advantageous. The great majority of radar-based hand gesture recognition systems use radar sensors with only a few antennas and focus on dynamic gestures~\cite{lit:RadarGestureTrajectory}. They often rely heavily on Doppler information from the sensor~\cite{lit:RadarGestureDoppler}.
By relying mostly on Doppler information, their use for static hand poses is very limited. This is because useful Doppler data can only be produced for objects in motion. In this work, static hand gestures—namely, the gestures for American Sign Language (ASL) alphabet—are classified using an imaging radar with 94 transmitting (TX) and 94 receiving (RX) antennas.
This allows the generation of an intensity image of the reflected object and also, to some extent, depth information from the received electromagnetic wave. Since the manual collection of measurement data is cumbersome and time-consuming, training data were generated with the avanced radar ray-tracing simulator for hand poses presented in~\cite{lit:braeunigRaytracingHand}, which is based on the implementation firstly proposed in~\cite{lit:schussler2021RealisticRadarsimulation}. One main goal of this work is to investigate how well a neural network (NN) can be trained solely on simulated data and still be useful for the evaluation on real measurements afterward. Two example images of the simulated and measured microwave images are depicted in Fig.~\ref{fig:firstImage}.

\section{Simulation Approach}
\begin{figure}
    \centering
    \includegraphics{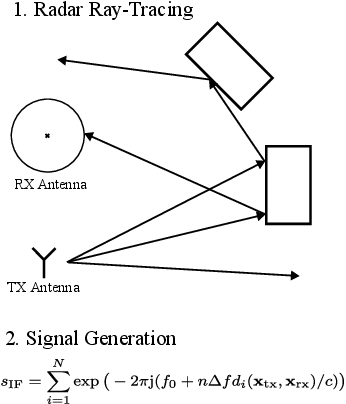}
    \caption{This figure shows the complete simulation process. In the first step, the scene is sampled by ray-tracing to collect all possible signal paths from all TX antennas to all RX antennas. In the second step, the final signal is generated.}
    \label{fig:SBRGO}
\end{figure}

The radar simulator is extensively described in the original publication by the authors~\cite{lit:schussler2021RealisticRadarsimulation}, with its application to hand poses investigated in the work proposed in ~\cite{lit:braeunigRaytracingHand}. This section will focus on the most important simulation steps. In the initial step, the scene is sampled with a large number of rays, launched from the TX positions in the simulation scene. These rays are then reflected on the hand's surface according to a user-defined material model, which will be explained later in this section.
Afterwards, each reflected ray that reaches an RX antenna within a specific radius (perception sphere) is considered in the signal generation step, which follows to generate the final signal. The concept can be likened to a shooting-and-bouncing-rays (SBR) approach~\cite{lit:orignalSBRPaper}, but instead of employing a Physical Optics (PO) approximation, it utilizes a simpler yet more flexible Geometrical Optics (GO) model. The simulator design is, therefore, closer to the concepts presented in~\cite{lit:Hirsenkorn} and~\cite{lit:BRDFRadar}. This process is depicted in Fig~\ref{fig:SBRGO}.

As previously discussed, rays are reflected according to a user-defined material model. This material model influences the directions in which rays are statistically reflected. The outgoing ray direction $ \mathbf{t_o}$ can be described according to the material model as
\begin{equation}
    \mathbf{t_o} = \alpha \mathbf{t_d} + (1-\alpha) \mathbf{t_s},
\end{equation}
where $\mathbf{t_d}$ is the direction for a completely diffuse reflection, and $\mathbf{t_s}$ is the ray direction for a pure specular reflection. The parameter $\alpha \in [0, 1]$ is a weighting factor that allows the model to behave in a more diffuse way (larger $\alpha$) or in a more specular way (smaller $\alpha$). The ray direction $\mathbf{t_d}$ can be computed using the following:
\begin{equation}
    \mathbf{t_d} = \mathbf{n} + \mathbf{r},
\end{equation}
where $\mathbf{n}$ is the normal vector of the surface, and $\mathbf{r}$ is a vector sampled randomly on a unit sphere. The specular direction $\mathbf{t_s}$ can be computed by
\begin{equation}
    \mathbf{t_s} = \mathbf{t_i} - 2 (\mathbf{t_i} \mathbf{n})\mathbf{n},
\end{equation}
with $\mathbf{t_i}$ being the incoming ray direction. In ~\cite{lit:braeunigRaytracingHand}, it was shown that realistic simulation results can be generated by low values for $\alpha$, simulating mostly specular reflection behavior. These findings were adapted in this work, and $\alpha$ was varied between the values 0.1 and 0.5 during the data generation process.

\subsection{Measurement Hardware and Signal Model}
\begin{figure}
    \centering
    \includegraphics[width=0.46\textwidth]{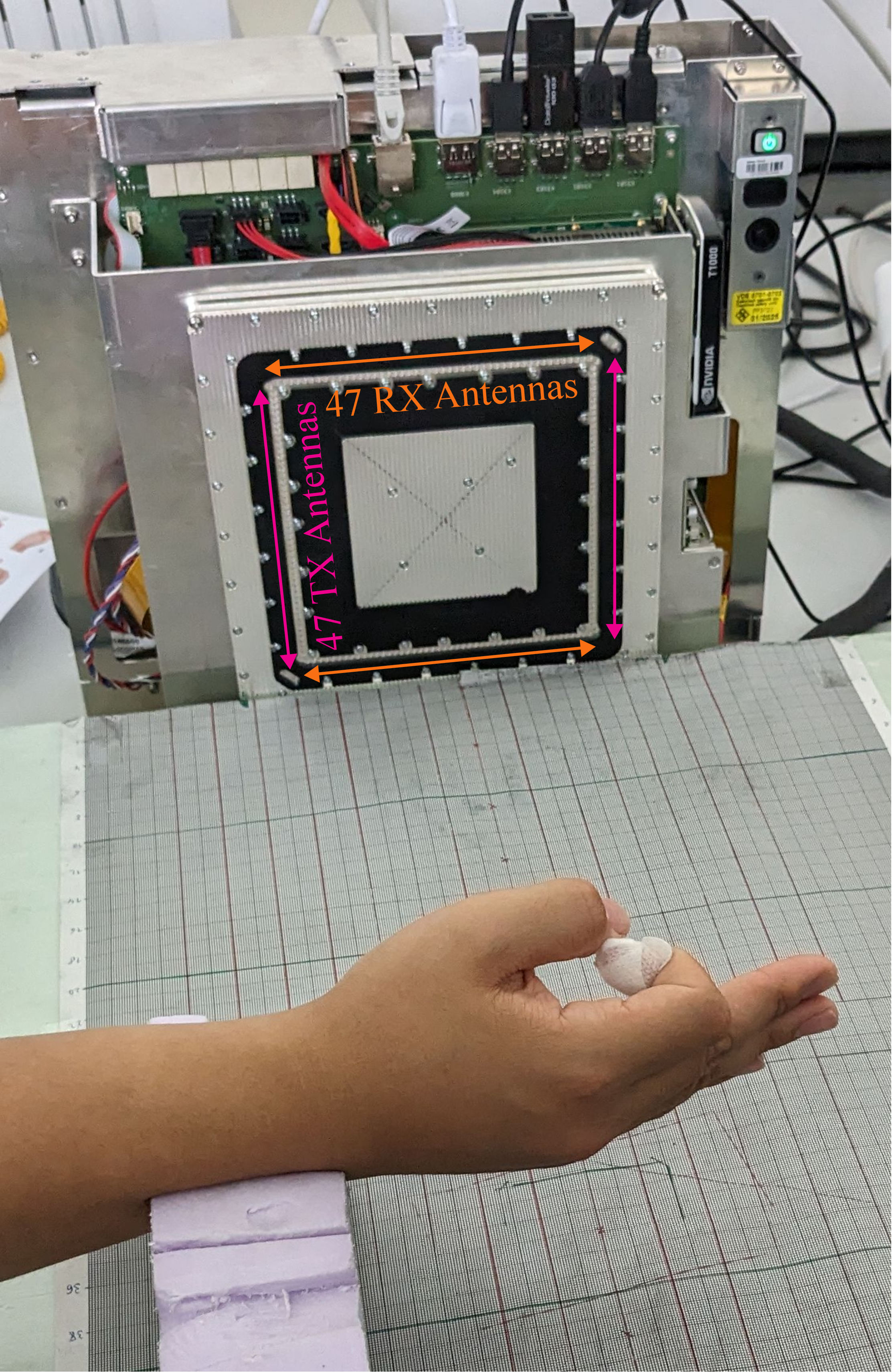}
    \caption{Photo of the measurement setup. The hand pose is sensed with the imaging radar in the back.}
    \label{fig:ImagingRadar}
\end{figure}

The measurement data is acquired by a stepped frequency continuous wave (SFCW) modulated radar operating between 72\,GHz and\,82 GHz. The intermediate frequency (IF) signal of this modulation scheme is stated below.
\begin{equation}
  s_{\textrm{IF}} (n, \mathbf{x}_\textrm{tx} , \mathbf{x}_\textrm{tx} ) = \sum_{i=1}^{N} \exp{ \big( -2 \pi \mathrm{j}(f_0 + n \Delta f d_i(\mathbf{x}_\textrm{tx}, \mathbf{x}_\textrm{rx} ) / c) \big)},
\end{equation}
whereby $\Delta f$ is the frequency step, and $f_0$ is the start frequency. In this application, a total of $N_\textrm{f} = 128$ frequency steps were used.Thus, 
the variable $n$ ranges from 0 to $N_\textrm{f} - 1$.In the equation above, the complete signal is a summation over each reflected signal for each of the $N$ reflection points denoted by the index $i$. The distance $d_i$ is the signal path length from each TX antenna $\mathbf{x}_\textrm{tx}$ to the scatter point $\mathbf{x}_i$, ending at the RX antenna position $\mathbf{x}_\textrm{rx}$.
\begin{equation}
    d_i = |\mathbf{x}_{tx} - \mathbf{x}_i| +  |\mathbf{x}_{\textrm{rx}} - \mathbf{x}_i|.
\end{equation}
The imaging radar device consists of 94 TX and 94 RX antennas arranged in a rectangular shape. The complete measurement setup, including the imaging radar, is depicted in Fig.~\ref{fig:ImagingRadar}.

After the IF signal was measured, the complete 3D volume of the region of interest is obtained by applying a matched-filter approach, as described, for example, in~\cite{lit:MicrowaveImaging}. Once the 3D volume is obtained, an orthogonal maximum projection along the depth axis is applied. This generates an intensity image with the strongest scattering points. At the same time, the depth value (z-coordinate) of the selected voxel is stored in a second image to generate the depth image of the measurement.

\subsection{3D Hand Model}
\begin{figure*}
    \centering
    \includegraphics[width=0.8\textwidth]{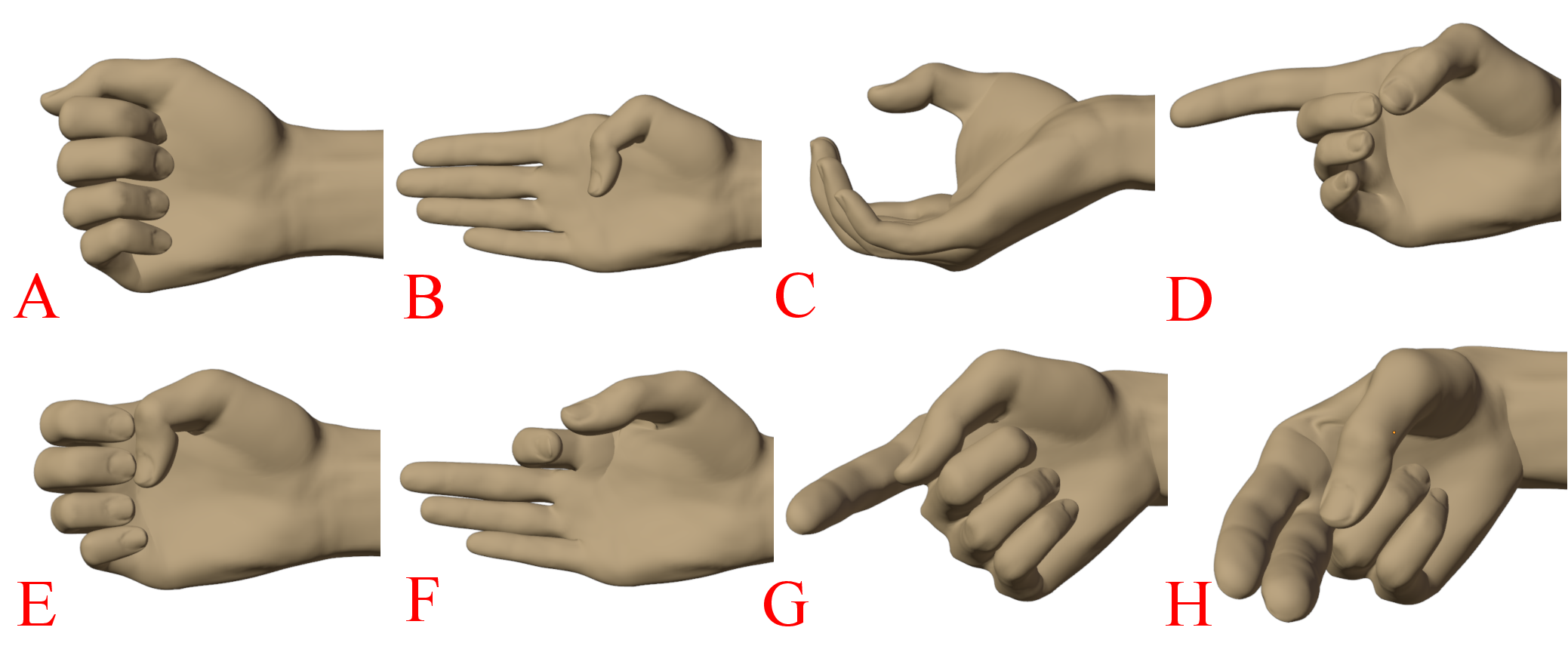}
    \caption{First 8 characters of the American Sign Language alphabet as 3D model.}
    \label{fig:FourHandPoses}
\end{figure*}

To simulate all hand poses, the hand model described in~\cite{lit:braeunigRaytracingHand} was used, and for the characters 'A' to 'H,' the related pose was formed by adjusting the bones fitted to the 3D mesh, as shown in Fig.~\ref{fig:FourHandPoses}.
The 3D models were obtained through photogrammetric reconstruction of three real human hands. Therefore, the overall shape of the models is very realistic, although they lack some details regarding skin wrinkles or other imperfections of human skin.

\section{Machine Learning Approach}
\begin{figure}[b]
    \centering
    \includegraphics[width=0.49\textwidth]{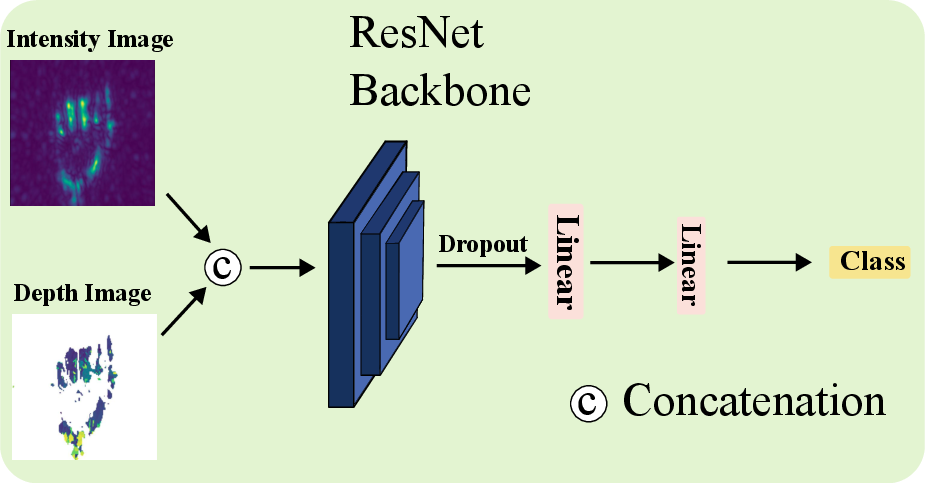}
    \caption{Complete workflow of the NN. Both intensity-image and depth-image are concatenated and fed into the ResNet backbone. A common classification head follows afterwards.}
    \label{fig:NeuralNetworkWorkflow}
\end{figure}

For this microwave image classification task, the popular ResNet~\cite{lit:He_2016_CVPR} network with varying numbers of residual layers (18, 34, 50, and 101) was chosen. The input of the NN is a concatenation of the intensity image and the depth image, which is then given into the backbone of the ResNet. Afterwards, the dimension is reduced to 8 classes using several linear layers interrupted by dropout layers. The complete workflow is depicted in Fig.~\ref{fig:NeuralNetworkWorkflow}.
Altogether, 7000 radar images were simulated with different viewing angles and various material parameter values. To achieve an even higher diversity of the data, the following data augmentation and pre-processing steps were applied to the images:
\begin{itemize}
    \item Random vertical and horizontal flips
    \item Rotation between 0 and 30 degrees
    \item Scaling the image size between 70\% and 100\% of its original size
    \item Convolution with a Gaussian kernel to prevent the network from overfitting to features that are specific to simulation.
\end{itemize}
Another employed regularization strategy is early stopping, which monitors validation loss during training. This techniques stops the training as soon as the improvement stagnates. 
This decision is based on a predefined 'patience' parameter for epochs without significant improvement on the validation set.

The NN was optimized with RAdam~\cite{lit:liu2019variance}, parameterized with a learning rate of 5e-5, reduced by a factor of 0.2 every 5 epochs. In total, the network was trained for 15 epochs with a batch size of 4.

\section{Results}
\begin{figure*}
    \centering
    \includegraphics[width=0.8\textwidth]{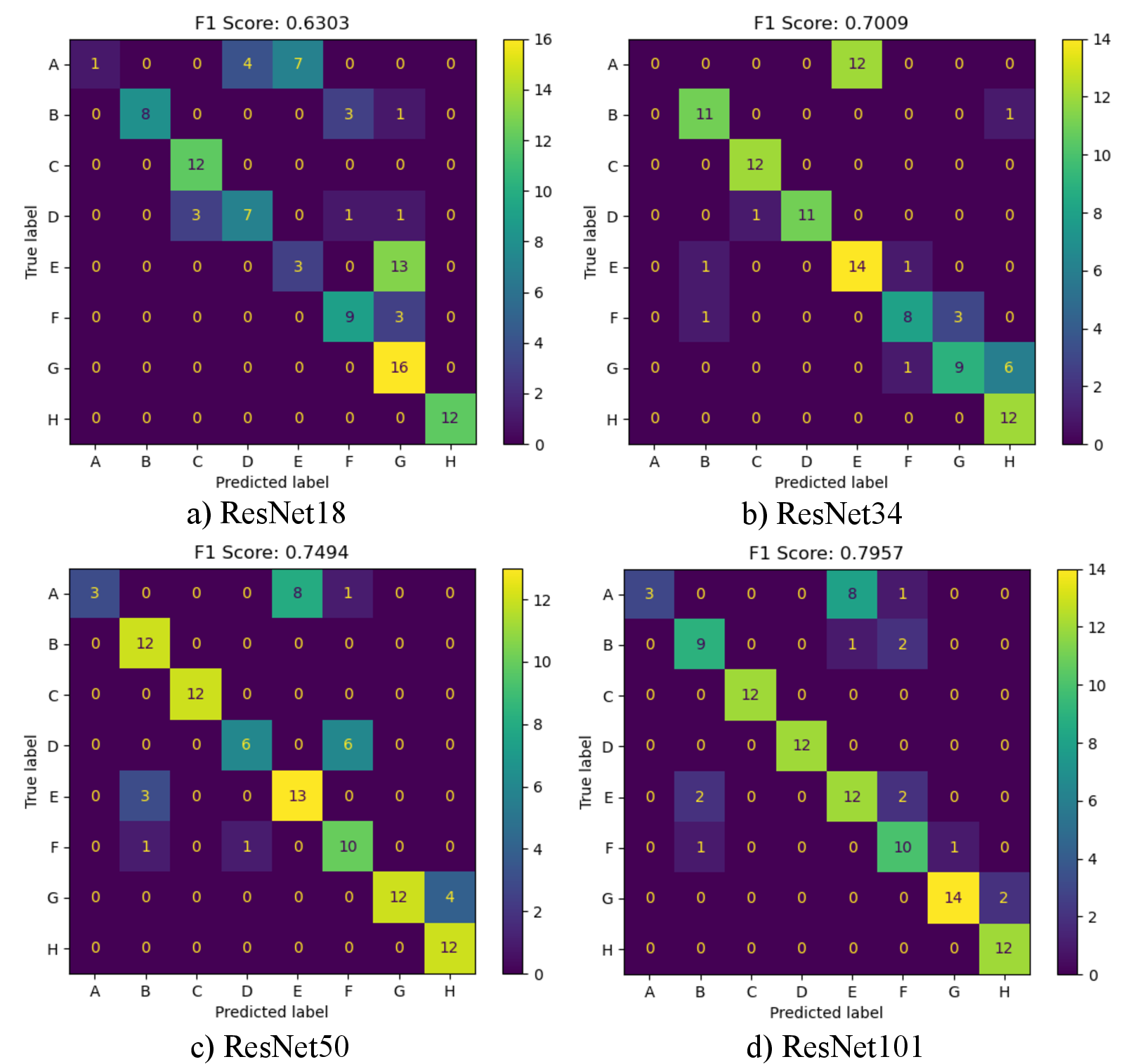}
    \caption{Test results of different ResNet backbones on the real-world dataset. Each image (a to d) is labeled with the corresponding ResNet model number.}
    \label{fig:validation_results}
\end{figure*}

The NN, trained solely on simulated data, was tested on 104 hand pose measurements. The resulting confusion matrices, along with the F1-scores, are depicted in Fig.~\ref{fig:validation_results}. Clearly, the performance increases with the model size, achieving the best score with 101 residual layers. However, even with 34 or 18 layers, the results are still satisfying. As can be seen, the classification of the character 'A' is the most challenging for the NN, often leading to confusion with the character 'E'. This is because both characters are essentially formed by making a fist-like pose. Comparing both hand signs, only the thumb has a different pose. Hence, if the thumb does not create a perceptible radar response due to unfavorable incident angles and specular reflections, a differentiation between both hand signs is challenging. All other characters were classified equally well.


\section{Discussion and Future Work}
This work proposes a neural network-based classifier for static hand signs of the American Sign Language alphabet using a microwave imaging radar. The study conclusively demonstrates that relying solely on simulated data is sufficient for training such classifiers and achieving robust classification results. Comparative analysis of different network sizes reveals the benefits of scaling neural networks for this problem. However, practical applications require a compromise between run-time, energy consumption, and classification performance.

In future research, exploring static hand poses, especially with smaller or sparse antenna arrays, could be valuable. Using fewer antenna channels can significantly reduce the costs and size of such systems, making them more affordable for consumer hardware in virtual reality applications. Reliable simulation software not only aids in generating training data and avoiding extensive measurement campaigns but also contributes to reducing development costs by creating a digital twin of the sensor before manufacturing a physical prototype.

\section*{Acknowledgment}
The authors would like to thank the Rohde \& Schwarz
GmbH \& Co. KG (Munich, Germany) for providing the radar
imaging devices and technical support that made this work
possible.

This work was partly funded by the Deutsche Forschungsgemeinschaft (DFG, German Research Foundation) – SFB 1483 – Project-ID 442419336, EmpkinS
%\IMSdisplayacksection{\IMSacktext}% end of \IMSdisplayacksection

%%%%%%%%%%%%%%%%%%%%%%%%%%%%%%%%%%%%%%%%%%%%%%%%%%%%%%%%%%%%%%%%%%%%%%%%%%%%%
\bibliographystyle{IEEEtran}
%\cleardoublepage
\bibliography{mybib}

%\begin{thebibliography}{00}
%    \bibitem{lit:ReviewAutomotiveSensors}F. Rosique, P. J. Navarro, C. Fernàndez, and A. Padilla, "A systematic review of perception system and simulators for autonomous vehicles research," Sensors, vol. 19, no. 3, p. 648, Feb. 2019, doi: 10.3390/s19030648.
%    \bibitem{lit:DeepLearningBasedHumanActivity}X. Li, Y. He, and X. Jing, “A survey of deep learning-based humanactivity recognition in radar,” Remote Sens., vol. 11, no. 9, 2019, Art. no. 1068.
%    \bibitem[lit:DinaK]{GaitMonitoring}
%\end{thebibliography}
\balance

\end{document}